\documentclass[runningheads]{llncs}
\usepackage[T1]{fontenc}
\usepackage{acro}
\usepackage{amsmath}
\usepackage{amssymb}
\usepackage{booktabs}
\usepackage{cleveref}
\usepackage{graphicx}
\usepackage{multirow}

\usepackage{xcolor}

\usepackage{soul}

\DeclareAcronym{name}{short = HashSCD, long = Hashing for Scene Change Detection}

\begin{document}
\title{From Image Hashing to Scene Change Detection}
\author{Anh-Kiet Duong\inst{1, 2}\orcidID{0009-0005-0230-6104} \and
Marie-Claire Iatrides\inst{1,3}\orcidID{0009-0005-3961-0564} \and
Petra Gomez-Krämer\inst{1}\orcidID{0000-0002-5515-7828} \and
Jean-Michel Carozza\inst{2}\orcidID{0000-0001-9077-9182}}
\authorrunning{Duong et al.}
\institute{L3i Laboratory, La Rochelle University, 17042 La Rochelle Cedex 1 - France\and
LIENSs Laboratory, La Rochelle University, 17042 La Rochelle Cedex 1 - France\and
Association Ferrocampus, 17100 Saintes - France
\email{\{anh.duong,marie-claire.iatrides,petra.gomez,jean-michel.carozza\}@univ-lr.fr}}
\maketitle              %
\begin{abstract}
Image hashing provides compact representations for efficient storage and retrieval but is inherently limited to global comparison and cannot reason about where changes occur. This limitation prevents hashing from being directly applicable to scene change detection, where spatial localization is essential. In this work, we revisit hashing from a scene change detection perspective and propose HashSCD, a patch-wise hashing framework that enables both efficient global change detection and localized change identification. HashSCD encodes spatially aligned patches into compact hash codes and aggregates them through an XOR-like operation, allowing change detection and localization to be performed directly in the Hamming space without repeated inference on previous images. The model is trained in an unsupervised manner using contrastive learning at both patch and global levels. Experiments demonstrate that HashSCD achieves competitive performance compared to state-of-the-art unsupervised hashing and scene change detection methods, while significantly reducing computational cost and storage requirements.

\keywords{Image hashing \and Scene change detection.}
\end{abstract}

\section{Introduction}
\label{sec:intro}

Scene change detection aims to identify significant differences in visual scenes observed at different time steps and plays a central role in long-term monitoring and surveillance applications~\cite{li2024umad}. Beyond determining whether a change has occurred, practical systems are often required to localize where the change happens while operating under strict constraints on computational costs and storage. These requirements become particularly critical in large-scale or long-term scenarios, where the same scene may be observed repeatedly over time~\cite{rs12101688}.

Most existing scene change detection methods rely on continuous feature representations extracted by deep neural networks~\cite{prabhakar2020cdnet++,li2020change}. While effective, such approaches typically require storing high-dimensional features or repeatedly forwarding previously observed images through the backbone network for comparison. This results in substantial computational overhead and memory consumption. As a result, it limits scalability in long-term monitoring settings. Moreover, many methods focus solely on dense change localization and do not support efficient retrieval of past observations, which is essential for temporal analysis and scene auditing.

In parallel, image hashing has been extensively studied as an efficient mechanism for compact representation, fast retrieval, and low-cost storage. By encoding images into short binary codes, hashing enables comparison in the Hamming space with near-constant time complexity~\cite{luo2023survey}. The Hamming space is a discrete binary space where distance is measured by the number of differing bits between two codes. However, existing hashing methods are primarily designed for global image retrieval and lack the ability to reason about spatially localized changes. As a result, conventional hashing is ill-suited for scene change detection, where spatial correspondence and localization are indispensable.

\begin{figure}
    \centering
    \includegraphics[width=0.9\textwidth]{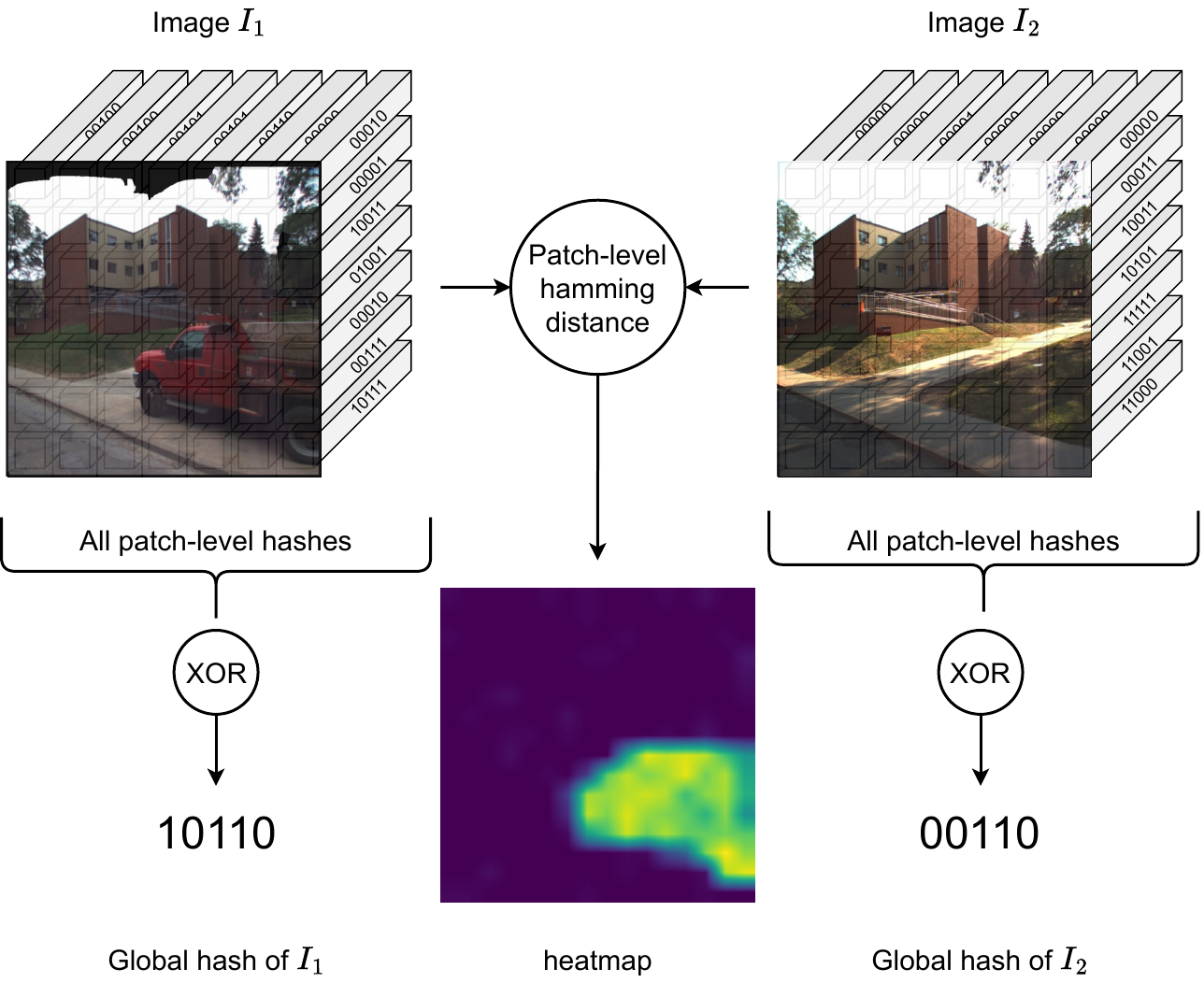}
    \caption{Overview of the proposed patch-wise hashing formulation for scene change detection, which enables efficient global comparison and localized change identification.}
    \label{fig:motiv}
\end{figure}

This work explores the question of whether hashing can be adapted to support scene change detection while retaining its inherent efficiency advantages. Specifically, we seek a representation that is compact enough for large-scale storage and fast comparison, yet expressive enough to localize changes without relying on dense supervision or repeated inference over past images. To this end, we propose a patch-wise hashing formulation that bridges global hashing efficiency and local change awareness within a unified framework, as illustrated in \cref{fig:motiv}. Our contributions are summarized as follows:
\begin{itemize}
    \item We introduce \ac{name}, the first hashing-based framework for scene change detection that supports both global comparison and localized change identification.
    \item We propose a patch-wise hashing and XOR-like aggregation scheme that enables efficient change localization while preserving the compactness required for hashing-based retrieval.
    \item We demonstrate that \ac{name} achieves competitive performance compared to state-of-the-art unsupervised hashing methods and unsupervised scene change detection approaches, while significantly reducing computational and storage costs.
\end{itemize}

\section{Related work}
\label{sec:related}
This section provides an overview of related work on unsupervised image hashing and scene change detection.

\subsection{Unsupervised image hashing}

Unsupervised image hashing has been extensively studied as a means of learning compact binary representations without labeled supervision. According to the taxonomy in~\cite{luo2023survey}, existing deep unsupervised hashing methods can be broadly categorized into three main groups. The first group relies on pseudo-label generation~\cite{hu2017pseudo,huang2016unsupervised,zhang2020deep}, where clustering or nearest-neighbor relations on pretrained features are used to construct surrogate supervision signals. The second group focuses on similarity reconstruction~\cite{tu2020mls3rduh,yang2018semantic,yang2019distillhash}, aiming to preserve pairwise similarities defined by distance metrics in the original feature space. The third group adopts prediction-free self-supervised learning strategies, such as autoencoder-based models~\cite{shen2020auto} or GAN-based frameworks~\cite{zieba2018bingan}, to learn binary codes without explicit similarity constraints.

More recently, contrastive learning has emerged as a dominant paradigm for unsupervised hashing~\cite{jang2021self,wang2022contrastive,ng2023unsupervised}, achieving state-of-the-art retrieval performance on generic benchmarks. While these methods perform well on wide-category datasets such as NUS-WIDE~\cite{chua2009nus} and CIFAR-10~\cite{krizhevsky2009learning}, their effectiveness significantly degrades on fine-grained datasets, including CUB200-2011~\cite{wah2011caltech} and Oxford Flowers~\cite{nilsback2008automated}, where subtle inter-class differences dominate.

The challenge of fine-grained image hashing has only recently been explicitly addressed. The work of~\cite{hu2024asymmetric} represents an early attempt in this direction, highlighting the limitations of existing unsupervised hashing methods under fine-grained settings. However, most current approaches still rely heavily on global representations, require multi-pass training, and remain susceptible to hash collisions, leading to increased computational cost and reduced discriminability. These limitations motivate the development of new hashing formulations that can effectively capture local discriminative patterns while maintaining the efficiency of compact binary representations.

\subsection{Scene change detection}

Scene change detection has been widely studied under supervised, unsupervised, and zero-shot settings. Supervised approaches typically rely on encoder-decoder architectures~\cite{thakur2025spectral} or continuous features extracted from shared CNN backbones to predict dense change masks~\cite{nguyen2018change,huang2020change}. While effective, these methods require pixel-level annotations and incur high computational costs, which limits their applicability in scenarios where labeled data is scarce.

To alleviate the dependence on annotation, unsupervised methods have been explored. For example, SSCD~\cite{ramkumar2021self} leverages continuous feature representations trained with self-supervised objectives such as Barlow Twins to generate change heatmaps. However, such approaches require storing high-dimensional continuous features and performing expensive distance computations, resulting in substantial memory and computational overhead. More recently, zero-shot methods exploited large pretrained models~\cite{cho2025zero}, such as Segment Anything Model (SAM)~\cite{kirillov2023segment}, to detect changes by comparing object-level segmentations before and after a scene change. Despite their flexibility, these methods struggle with changes that are not object-centric, such as structural damage or texture-level variations, as observed in the TSUNAMI subset of the PCD dataset \cite{jst2015change}.

These limitations raise an important question: can scene changes be detected efficiently without dense supervision, while significantly reducing computational and storage costs? This motivates the exploration of hashing-based representations, which enable compact storage and efficient comparison, yet remain largely unexplored in the context of scene change detection.

\section{Proposed method}
\label{sec:method}
\begin{figure}[ht]
    \centering
    \includegraphics[width=\linewidth]{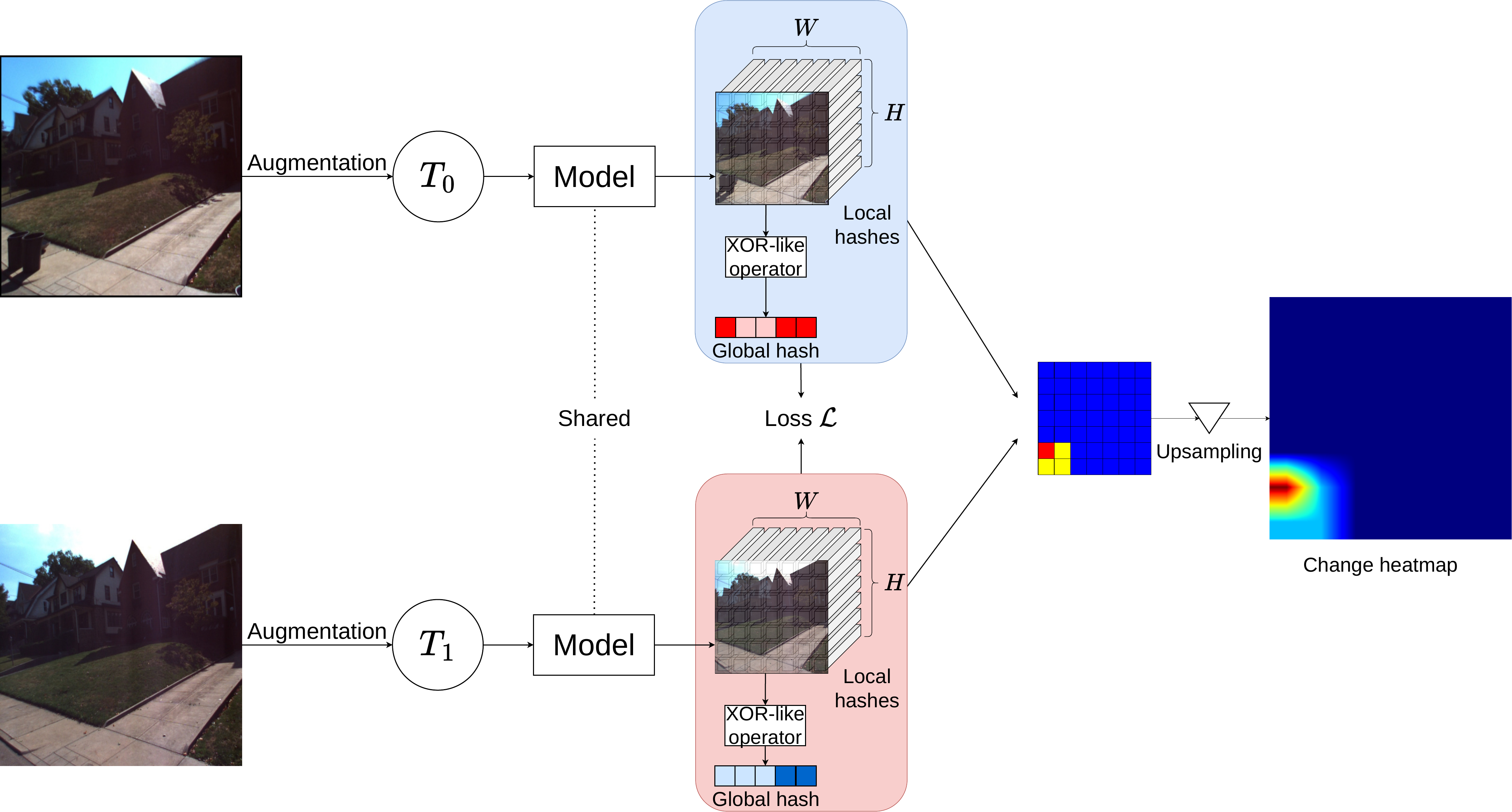}
    \caption{Overview of the proposed \ac{name} framework for scene change detection. Given two images captured at times $T_0$ and $T_1$, augmented views are processed by a shared backbone to extract patch-wise features, which are mapped to local hash embeddings. An XOR-like aggregation is applied to the local hashes to obtain a compact global hash representation for retrieval. During training, contrastive loss is imposed at both patch and global levels. During inference, global hashes are compared using the Hamming distance to determine whether a change occurs, while patch-wise hash differences are used to generate a change heatmap via upsampling.
    }
    \label{fig:framework}
\end{figure}

In this section, we present our \ac{name}, a framework for scene change detection that jointly supports global comparison and localized change identification. An overview of the proposed framework is illustrated in \cref{fig:framework}. We first introduce a patch-wise hash embedding and a recursive aggregation scheme to construct compact global representations while preserving spatial information (\cref{sec:patch}). We then describe a soft hashing strategy and an unsupervised contrastive optimization objective to enable end-to-end training in the Hamming space (\cref{sec:contrastive}). Finally, we detail how the learned hash codes are used for efficient change detection, localization, and long-term storage across multiple temporal observations (\cref{sec:scd}).

\subsection{Patch-wise hash embedding}
\label{sec:patch}
In scene change detection, capturing spatially localized differences between scenes is crucial. Representing an entire image with a single global hash inevitably discards spatial correspondence and prevents reasoning about where the changes occur. To address this limitation, we introduce a patch-wise image hashing method that uses a set of patch-wise hash embeddings, each corresponding to a fixed spatial location.

Given an input image $x$, a backbone network produces a feature map 
$\mathbf{F} \in \mathbb{R}^{C \times H \times W}$, where $C$, $H$, and $W$ denote the channel depth, height, and width of the backbone output, respectively. The feature map is broken down into $P = H \times W$ local feature vectors $\{\mathbf{f}_i\}_{i=1}^{P}$, where $\mathbf{f}_i \in \mathbb{R}^{C}$. For each spatial location, a compact hash representation is obtained by applying a hashing projection to the corresponding local feature:
\begin{equation}
\mathbf{h}_i = \phi(\mathbf{\mathcal{W}}\mathbf{f}_i),
\end{equation}
where $\mathbf{\mathcal{W}} \in \mathbb{R}^{l \times C}$ denotes the learnable hashing projection and $l$ is the hash length. The mapping function $\phi(\cdot)$ transforms real-valued projections into the hash space. Its exact formulation is described in \cref{sec:contrastive}.

To obtain a single hash representation for the image while retaining contributions from all local regions, we aggregate the patch-wise hashes through a sequential absolute-difference operation:

\begin{equation}
\label{eq:xor}
    \mathbf{R} = \left | \mathbf{h}_{P}-\left | \mathbf{h}_{P-1} - \cdots  \left | \mathbf{h}_4 -   \left | \mathbf{h}_3 - \left | \mathbf{h}_2 - \mathbf{h}_1 \right | \right | \right | \cdots   \right | \right |
\end{equation}

The final aggregated hash for the image is given by $\mathbf{R}$. By iterating over patch-wise hashes from $1$ to $P$, this operation combines all patch-wise hashes into a unified representation $\mathbf{R}$ without collapsing them through averaging or pooling. Since each patch-wise hash $\mathbf{h}_i \in \mathbb{R}^l$, the aggregated hash $\mathbf{R}$ also lies in $\mathbb{R}^l$, preserving a fixed hash length regardless of the number of patches. When operating in the binary domain, the absolute-difference aggregation is equivalent to a cascaded bitwise XOR as proven in Appendix \ref{sec:proof}, yielding a compact global hash that encodes the joint configuration of local patterns.

This aggregation plays a dual role in the proposed framework. On one hand, by constructing the global hash from spatially distributed patch-wise hashes, it overcomes the limitation of conventional global hashing methods that are unable to reflect localized changes. On the other hand, it preserves the compactness required for efficient hashing-based retrieval by producing a single global hash through XOR-style aggregation, rather than concatenating all local hash codes into a high-dimensional representation. As a result, the proposed design bridges global hashing efficiency and local change awareness.

\subsection{Soft hashing and contrastive optimization}
\label{sec:contrastive}

The patch-wise hash embedding introduced in the previous subsection relies on discrete binary operations. However, directly learning binary hash codes is not feasible with gradient-based optimization due to the non-differentiability of hard binarization. To address this issue, we adopt a soft hashing strategy during training~\cite{cao2017hashnet}, which allows the hash codes to gradually approach binary values while remaining differentiable.

Specifically, given a real-valued representation $\mathbf{v}_i$ extracted from the backbone network, we obtain a soft hash embedding by
\begin{equation}
\mathbf{u}_i = \tanh(\mathbf{v}_i),
\end{equation}
where $\mathbf{u}_i \in (-1, 1)^l$ serves as a continuous relaxation of the binary hash code of length $l$. During inference, the soft representation is converted into a binary code using the sign function:
\begin{equation}
\mathbf{u} = \operatorname{sgn}(\mathbf{v}), \quad
\operatorname{sgn}(x) =
\begin{cases}
1, & x \geq 0, \\
-1, & x < 0.
\end{cases}
\end{equation}

In many existing hashing works, binary codes are represented in $\{-1,1\}$, where $-1$ is treated as bit $0$ and $1$ corresponds to bit $1$. In our formulation, the hash values are treated as continuous during training to enable gradient-based optimization. Specifically, the recursive aggregation in \cref{eq:xor} operates on soft hash codes in $[0,1]$, obtained via
\begin{equation}
\phi(\cdot) = \frac{\tanh(\cdot) + 1}{2}.
\end{equation}
During inference and storage, the soft hash codes are binarized to $\{0,1\}$ and stored as compact binary bits, ensuring efficient memory usage while preserving compatibility with Hamming distance computation.

To train the hashing network in an unsupervised manner, we adopt a contrastive learning objective inspired by asymmetric self-supervised hashing~\cite{hu2024asymmetric}. For each image, we construct an anchor view $(a)$ and a positive view $(p)$ corresponding to the same scene at times $T_0$ and $T_1$, respectively, augmented from the same image. The negative views $(n)$ are formed by combining anchor and positive views from the remaining $N-1$ samples in the minibatch, resulting in $2(N-1)$ negative views. The augmentations can be found in the Appendix~\ref{sec:addexp}.

Given the hash representations $\mathbf{u}^{(a)}$, $\mathbf{u}^{(p)}$, and $\mathbf{u}^{(n)}$, the asymmetric contrastive loss is defined as
\begin{equation}
\mathcal{L}_{\text{contr}}
=
- \log
\frac{
\exp \left( \mathbf{u}^{(a)} \cdot \mathbf{u}^{(p)} / \tau \right)
}{
\exp \left( \mathbf{u}^{(a)} \cdot \mathbf{u}^{(p)} / \tau \right)
+ \sum_{j=1}^{2(N-1)} \exp \left( \mathbf{u}^{(a)} \cdot \mathbf{u}^{(n)}_j / \tau \right)
},
\end{equation}
where $\tau$ denotes a temperature hyper-parameter.

In our framework, the asymmetric contrastive loss is applied on two levels. Firstly, it is implemented for each patch-wise hash embedding to encourage local consistency across time while preserving discriminative spatial patterns. Secondly, it is applied to the aggregated global hash obtained from the patch-wise aggregation. Since the global hash $\mathbf{R}$ lies in $[0,1]^b$, we map it back to a symmetric space via $2\mathbf{R} - 1$ before contrastive optimization.

The final objective combines the patch-level and global-level losses as
\begin{equation}
\mathcal{L}
=
\frac{1}{P+1} \left( \mathcal{L}_{\text{global}}
+ \sum_{i=1}^{P} \mathcal{L}_{\text{patch}}^{(i)} \right),
\end{equation}
where $\mathcal{L}_{\text{global}}$ denotes the asymmetric contrastive loss applied to the global hash and $\mathcal{L}_{\text{patch}}^{(i)}$ corresponds to the asymmetric contrastive loss computed for the $i$-th patch-wise hash.

\subsection{Hash-based change detection}
\label{sec:scd}
The patch-wise hash aggregation produces a set of local discrepancy outputs defined on a coarse spatial grid. Specifically, patch-level distances are computed by comparing hash embeddings at corresponding spatial locations, yielding a change response for each position. To obtain a dense change representation aligned with the input resolution, we upsample these patch-level distances $\mathbf{R}$ using interpolation guided by the spatial layout of the CNN feature maps. This process preserves spatial correspondence while avoiding learnable parameters.

Beyond change localization, the proposed hashing formulation supports long-term scene monitoring across multiple time steps. For a fixed observation location, we store both the aggregated global hash and patch-wise hash embeddings corresponding to different temporal states. Given a newly observed image, its global hash can be directly compared with stored global hashes using the Hamming distance to determine whether a scene change has occurred. Once a discrepancy is detected at the global level, the stored patch-wise hashes are used to localize the change by comparing local hash codes at corresponding spatial positions. Importantly, this design avoids re-processing previously observed images through the backbone network, significantly reducing computational overhead. Since each state image is represented by $(P+1) \times l$ bits for patch-wise and global hashes, the storage cost remains compact for long-term monitoring.

All comparisons and storage operations are performed in the hash space. Instead of storing high-dimensional floating-point features, the system maintains compact binary hash codes of length $l$. As a result, distance computation reduces to the Hamming distance, which is more efficient than other distances in both memory and computation. During inference, the recursive aggregation in \cref{eq:xor} simplifies into a sequence of bitwise XOR operations between binary hash codes, further lowering computational cost. This design enables scalable change detection and localization in long-term, large-scale monitoring scenarios.

\section{Experiments}
\label{sec:experiments}
In this section, we evaluate the proposed \ac{name} introduced in \cref{sec:method}. We first describe the experimental setup and implementation details, followed by quantitative evaluations on scene change detection and retrieval benchmarks. We then analyze the impact of key design choices through ablation studies and present qualitative results to illustrate the behavior of the proposed method.

\subsection{Setup and datasets}

We implement \ac{name} in PyTorch~\cite{paszke2019pytorch} and conduct all experiments on an NVIDIA H100 GPU. Following common practice in unsupervised hashing~\cite{hu2024asymmetric}, we adopt VGG16~\cite{simonyan2014very} as the backbone network. The model is trained for $100$ epochs with a batch size of $64$ using the Adam optimizer~\cite{kingma2014adam}. The learning rate is set to $10^{-5}$ for the backbone and $10^{-2}$ for the remaining layers, consistent with prior work~\cite{wang2022contrastive}. The temperature hyper-parameter $\tau$ is fixed to $0.3$ following~\cite{hu2024asymmetric}. Following prior work~\cite{hu2024asymmetric,ramkumar2021self}, we evaluate the retrieval performance using mean Average Precision (mAP) and assess change detection results using the F1 score at a threshold of $0.5$, where a pixel is classified as changed if its normalized Hamming distance after upsampling exceeds $0.5$.

The VL-CMU-CD dataset~\cite{alcantarilla2018street} consists of $1{,}362$ RGB images captured from $152$ viewpoints in urban environments. It mainly reflects macroscopic scene changes and is characterized by significant noise, which often obscures fine-grained differences. The Panoramic Change Detection (PCD) dataset~\cite{jst2015change} includes two subsets, namely TSUNAMI and GSV, each containing $100$ pairs of panoramic images. The TSUNAMI subset depicts post-disaster scenes in tsunami-affected regions of Japan, while the GSV subset consists of image pairs from Google Street View. For both datasets, images are captured at two different time steps, denoted as $T_0$ and $T_1$, along with corresponding change masks indicating regions of change. Compared to VL-CMU-CD, PCD images exhibit fewer noisy changes but involve larger viewpoint variations.

For retrieval experiments, we evaluate on three standard benchmarks. Oxford Flowers~\cite{nilsback2008automated} contains $8{,}189$ images from $102$ flower categories and represents a small-scale fine-grained retrieval setting. Food101~\cite{bossard2014food} includes $101{,}000$ images across $101$ food categories with higher intra-class variation represents a large-scale fine-grained retrieval setting. NUS-WIDE~\cite{chua2009nus} consists of $159{,}143$ images annotated with $21$ semantic concepts in a multi-label setting, representing a large-scale and wide-category retrieval benchmark.

\subsection{Change detection}
\label{sec:change_exp}

We evaluate \ac{name} on scene change detection benchmarks and compare it with state-of-the-art continuous-feature-based methods, including the supervised method C-3PO~\cite{wang2023reduce}, the zero-shot method ZSSCD~\cite{cho2025zero}, and the unsupervised methods SSCD and D-SSCD~\cite{ramkumar2021self}. Experiments are conducted on the VL-CMU-CD~\cite{alcantarilla2018street} and PCD~\cite{jst2015change} datasets, which consist of paired images captured at the same location across different time steps. For \ac{name}, we evaluate different hash lengths $l$ for the patch-wise hashes to analyze their impact on change detection performance. Quantitative results are reported in \cref{tab:change}.

\begin{table}[ht]
\centering
\caption{Change detection performance (F1 score) on VL-CMU-CD and PCD datasets.}
\label{tab:change}
\begin{tabular}{lccc}
\hline
Method                                & Bits       & VL-CMU-CD      & PCD            \\ \hline
\multicolumn{4}{l}{\textit{Supervised}}                                              \\
\multicolumn{2}{l}{C-3PO \cite{wang2023reduce}}    & 0.794          & 0.824          \\ \hline
\multicolumn{4}{l}{\textit{Zero-shot}}                                               \\
\multicolumn{2}{l}{ZSSCD \cite{cho2025zero}}       & 0.516          & 0.565          \\ \hline
\multicolumn{4}{l}{\textit{Unsupervised (continuous features)}}                      \\
\multicolumn{2}{l}{SSCD \cite{ramkumar2021self}}   & 0.745          & 0.583          \\
\multicolumn{2}{l}{D-SSCD \cite{ramkumar2021self}} & 0.725          & 0.642          \\ \hline
\multicolumn{4}{l}{\textit{Unsupervised (hashing)}}                                  \\
\multirow{3}{*}{Our \ac{name}}        & 16         & 0.641          & 0.635          \\
                                      & 32         & 0.681          & 0.648          \\
                                      & 64         & \textbf{0.694} & \textbf{0.695} \\ \hline
\end{tabular}
\end{table}

Beyond detection accuracy, we emphasize the computational and storage efficiency of the proposed hashing-based formulation. Unlike continuous representations that rely on high-dimensional floating-point features, \ac{name} operates on compact binary hash codes. For a VGG16 backbone, continuous features typically have a dimensionality of 512 floats, whereas our hash representation uses only 16, 32, or 64 bits. This difference yields a substantial reduction in storage cost: for a $512 \times 512$ image, storing continuous features on a $32 \times 32$ spatial grid requires $32 \times 32 \times 512 \times 32$ bits, corresponding to about $2$ MB per image, while hashing with $64$-bit codes requires only $32 \times 32 \times 64$ bits, about $8$ kB per image, yielding a $256$-fold reduction. In addition, distance computation is performed in the Hamming space rather than the continuous feature space, resulting in a comparable $256$-fold reduction in computational complexity.

In addition to change detection, we further evaluate the retrieval capability of \ac{name} within the scene change detection setting. As reported in \cref{tab:retrival_change}, images captured at time $T_1$ are used as queries to retrieve their corresponding images at time $T_0$ from the same location. This setting reflects a practical scenario where historical observations are retrieved to assess temporal changes without re-processing previously stored images.

\begin{table}[ht]
\centering
\caption{Retrieval performance (mAP) for retrieving $T_0$ images using $T_1$ queries.}
\label{tab:retrival_change}
\begin{tabular}{l|ccc|ccc}
\hline
\multirow{2}{*}{Methods} & \multicolumn{3}{c|}{VL-CMU-CD} & \multicolumn{3}{c}{PCD} \\
                         & 16       & 32       & 64       & 16     & 32     & 64    \\ \hline
Our \ac{name}            & 0.155    & 0.386    & 0.660    & 0.336  & 0.550  & 0.933 \\ \hline
\end{tabular}
\end{table}

As shown in the results, \ac{name} achieves competitive performance compared to supervised and unsupervised continuous-feature-based methods, while outperforming zero-shot approaches. On the VL-CMU-CD dataset, a modest performance gap relative to continuous-feature-based methods is observed, which is expected since they operate on high-dimensional representations that preserve richer appearance information. Notably, \ac{name} is the first hashing-based framework for scene change detection, enabling both retrieval and change localization without requiring an additional forward pass through the backbone of the previous image. By operating entirely in the hash space, the proposed method significantly reduces computational cost and storage requirements, making it well-suited for long-term and large-scale scene monitoring.

\subsection{Retrieval performance}
As discussed in \cref{sec:change_exp}, the proposed \ac{name} supports retrieval through its global hash representation and has been evaluated on scene change detection benchmarks. In this subsection, we extend the evaluation to standard image retrieval benchmarks and compare it with a range of unsupervised hashing methods, including Unsupervised Greedy Hash (UGH)~\cite{su2018greedy}, MLS$^3$RUDH~\cite{tu2020mls3rduh}, Bihalf \cite{li2021deep}, CIBHash~\cite{qiu2021unsupervised}, SPQ~\cite{jang2021self}, MeCoQ~\cite{wang2022contrastive}, SDC~\cite{ng2023unsupervised}, and $A^2$-SSL~\cite{hu2024asymmetric}. Most baseline results are taken from the original papers, and we use official implementations when available. Retrieval performance is reported on three datasets with increasing scale and diversity, ranging from the small-scale Oxford Flowers~\cite{nilsback2008automated} to the large-scale Food101~\cite{bossard2014food} and NUS-WIDE~\cite{chua2009nus}. These datasets cover both fine-grained and wide-category retrieval settings, where recent hashing methods are known to struggle particularly on fine-grained data~\cite{hu2024asymmetric}. For a fair comparison with prior work, all images are resized to $224\times224$, and hash lengths are set to $\{12, 32, 48\}$ for fine-grained datasets and $\{16, 32, 64\}$ for NUS-WIDE.

\begin{table}[ht]
\caption{Retrieval performance (mAP) comparison on standard image retrieval benchmarks.}
\centering
\label{tab:retrieval}
\begin{tabular}{l|ccc|ccc|ccc}
\hline
\multirow{2}{*}{Methods} & \multicolumn{3}{c|}{Oxford Flowers} & \multicolumn{3}{c|}{Food101} & \multicolumn{3}{c}{NUS-WIDE} \\
 & 12 & 32 & 48 & 12 & 32 & 48 & 16 & 32 & 64 \\ \hline
UGH \cite{su2018greedy} & 0.073 & 0.121 & 0.124 & 0.043 & 0.073 & 0.084 & 0.633 & 0.691 & 0.731 \\
ML$S^3$RUDH \cite{tu2020mls3rduh} & 0.116 & 0.189 & 0.204 & 0.051 & 0.086 & 0.096 & 0.713 & 0.727 & 0.750 \\
Bihalf \cite{li2021deep} & 0.136 & 0.192 & 0.212 & 0.062 & 0.092 & 0.102 & 0.774 & 0.801 & 0.819 \\
CIBHash \cite{qiu2021unsupervised} & 0.185 & 0.298 & 0.330 & 0.073 & 0.117 & 0.122 & 0.771 & 0.797 & 0.809 \\
SPQ \cite{jang2021self} & 0.203 & 0.301 & 0.349 & 0.080 & 0.122 & 0.133 & - & - & - \\
MeCoQ \cite{wang2022contrastive} & 0.191 & 0.303 & 0.344 & 0.079 & 0.116 & 0.132 & 0.802 & 0.822 & 0.832 \\
SDC \cite{ng2023unsupervised} & 0.179 & 0.261 & 0.284 & 0.075 & 0.104 & 0.116 & \textbf{0.807} & \textbf{0.823} & \textbf{0.834} \\
$A^2$-SSL \cite{hu2024asymmetric} & 0.342 & 0.448 & 0.464 & 0.085 & 0.141 & 0.154 & - & - & - \\ \hline
Our \ac{name} & \textbf{0.349} & \textbf{0.453} & \textbf{0.484} & \textbf{0.092} & \textbf{0.150} & \textbf{0.156} & 0.799 & \textbf{0.823} & 0.829 \\ \hline
\end{tabular}
\end{table}

As shown in \cref{tab:retrieval}, \ac{name} consistently outperforms state-of-the-art methods on fine-grained datasets, thanks to patch-wise representations that capture subtle local differences. On wide-category datasets, the gain is less pronounced, as the method relies on patch-wise backbone features without directly leveraging the pretrained global projector, which is particularly effective for wide-category recognition. Nevertheless, our method remains competitive with existing state-of-the-art hashing approaches. Overall, these results show that the proposed \ac{name} generalizes well across different retrieval regimes.

\subsection{Ablation studies}

We perform ablation studies to examine the effect of key design choices in \ac{name}, with results reported in \cref{tab:ablation}. Removing the global contrastive loss $\mathcal{L}_{\text{global}}$ leads to consistent performance drops, indicating its importance for learning discriminative global representations. The temperature parameter $\tau$ also influences the performance, where $\tau=0.2$ yields slightly better results for change detection. Regarding backbone architectures, ResNet50~\cite{he2016deep} pretrained on ImageNet shows similar change detection accuracy, while DINOv2 ViT-B/14~\cite{oquab2023dinov2} achieves improved performance, which can be attributed to its finer patch partitioning that enables more accurate localization of scene changes.

\begin{table}[ht]
\centering
\caption{Ablation study on key design choices in HashSCD.}
\label{tab:ablation}
\begin{tabular}{l|cc|cc}
\hline
\multirow{3}{*}{Setting}          & \multicolumn{2}{c|}{Retrieval}      & \multicolumn{2}{c}{Change detection} \\
                                  & \multicolumn{2}{c|}{Oxford Flowers} & \multicolumn{2}{c}{VL-CMU-CD}        \\
                                  & 12               & 48               & 16                & 64               \\ \hline
w/o $\mathcal{L}_{\text{global}}$ & 0.339            & 0.482            & 0.634             & 0.691            \\
$\tau=0.4$                        & 0.332            & 0.472            & 0.629             & 0.657            \\
$\tau=0.2$                        & 0.337            & 0.499            & 0.638             & 0.705            \\
ResNet50 ImageNet                 & 0.302            & 0.550            & 0.614             & 0.689            \\
DINOv2 ViT-B/14                   & \textbf{0.349}   & \textbf{0.677}   & \textbf{0.651}    & \textbf{0.727}   \\ \hline
Our \ac{name}                     & \textbf{0.349}   & 0.484            & 0.641             & 0.694            \\ \hline
\end{tabular}
\end{table}

\subsection{Qualitative results}

\begin{figure}[h]
\centering
\newcommand{\ncols}{5}
\setlength{\tabcolsep}{0.1em}

\newlength{\imgwidth}
\setlength{\imgwidth}{\dimexpr(0.9\textwidth - \ncols\tabcolsep)/\ncols\relax}

\begin{tabular}{*{\ncols}{c}}
\includegraphics[width=\imgwidth]{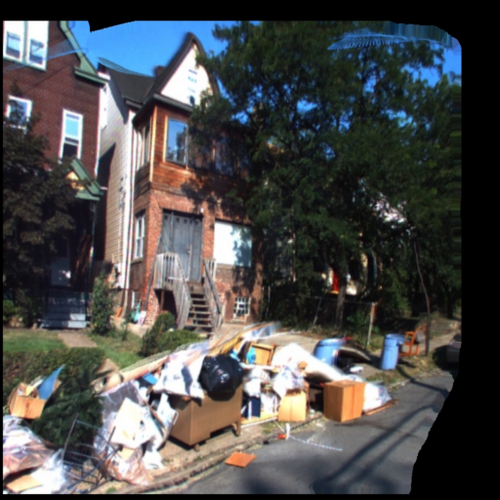} &
\includegraphics[width=\imgwidth]{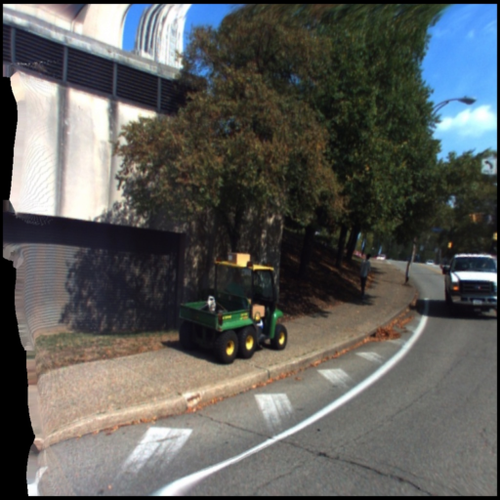} &
\includegraphics[width=\imgwidth]{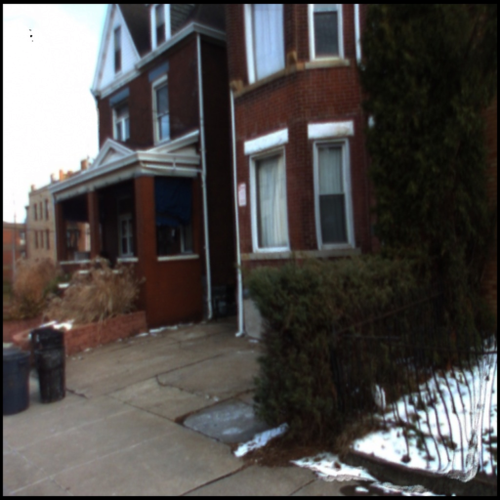} &
\includegraphics[width=\imgwidth]{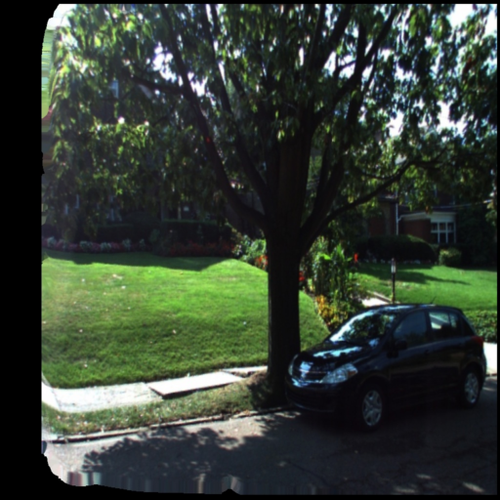} &
\includegraphics[width=\imgwidth]{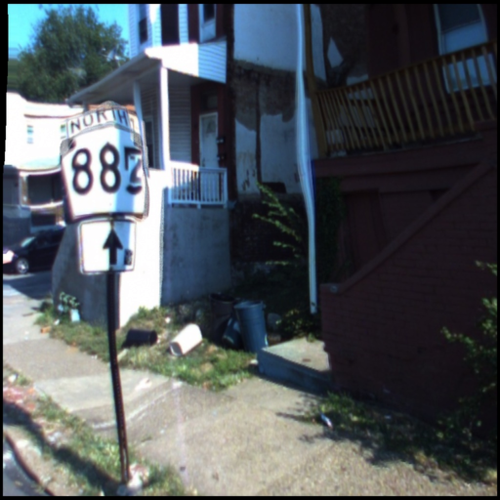} \\[-0.3em]

\multicolumn{5}{c}{$T_0$ images} \\

\includegraphics[width=\imgwidth]{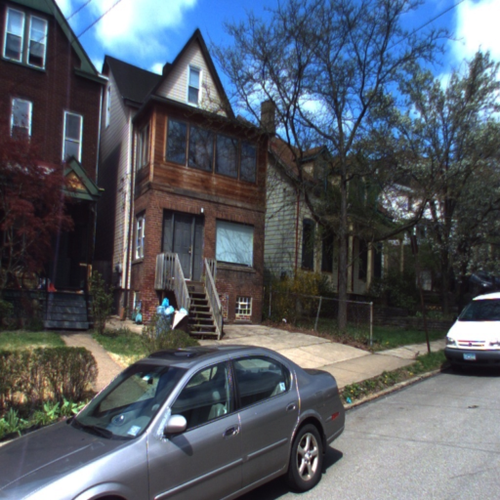} &
\includegraphics[width=\imgwidth]{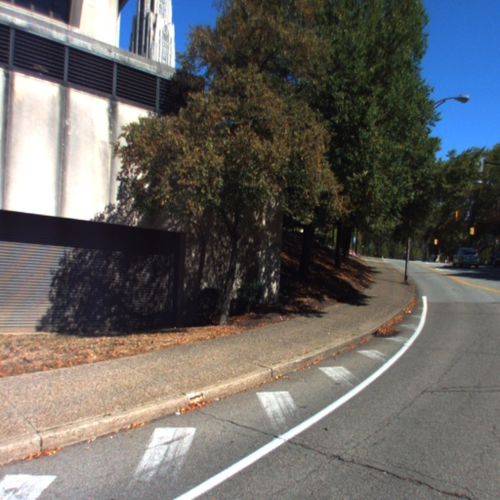} &
\includegraphics[width=\imgwidth]{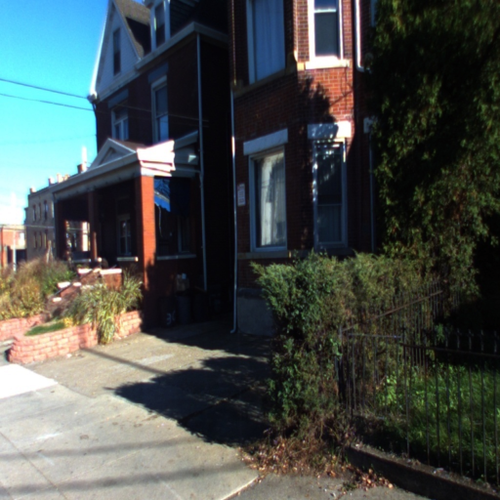} &
\includegraphics[width=\imgwidth]{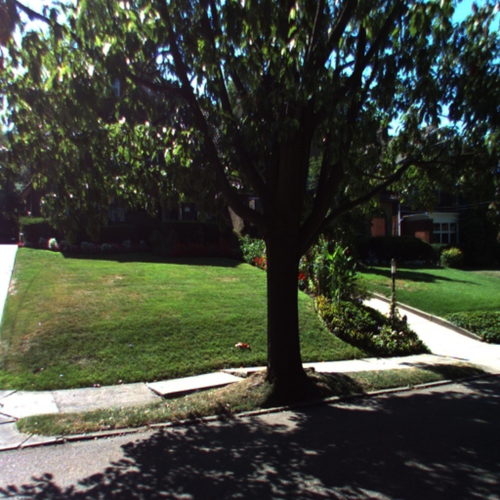} &
\includegraphics[width=\imgwidth]{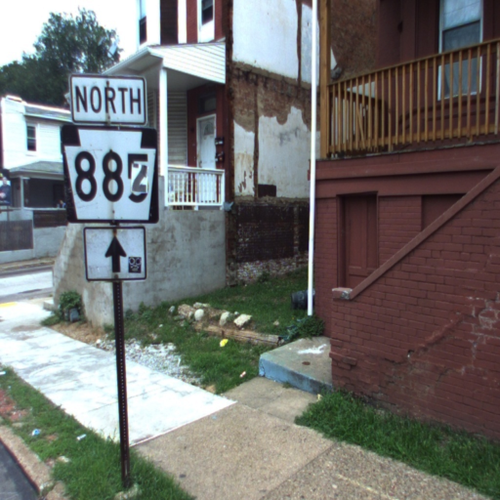} \\[-0.3em]

\multicolumn{5}{c}{$T_1$ images} \\

\includegraphics[width=\imgwidth]{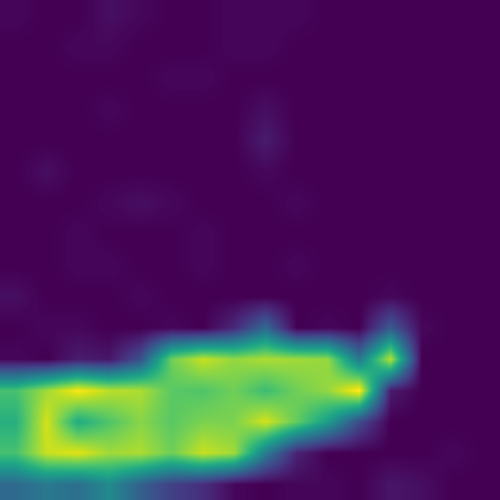} &
\includegraphics[width=\imgwidth]{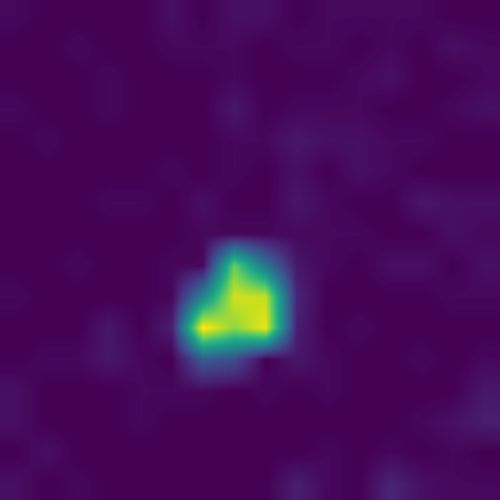} &
\includegraphics[width=\imgwidth]{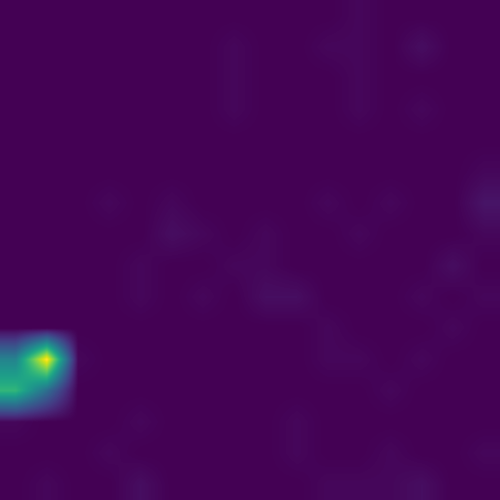} &
\includegraphics[width=\imgwidth]{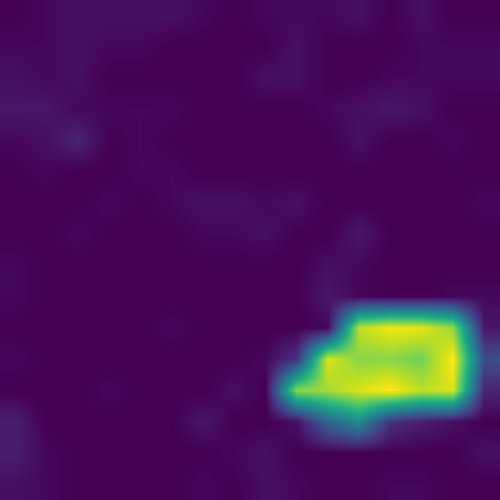} &
\includegraphics[width=\imgwidth]{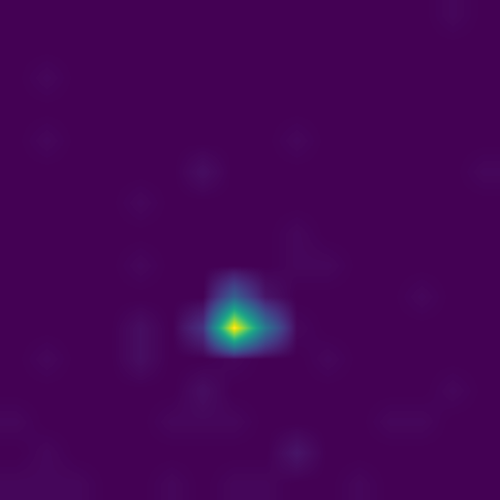} \\[-0.3em]

\multicolumn{5}{c}{Output heatmaps} \\

\includegraphics[width=\imgwidth]{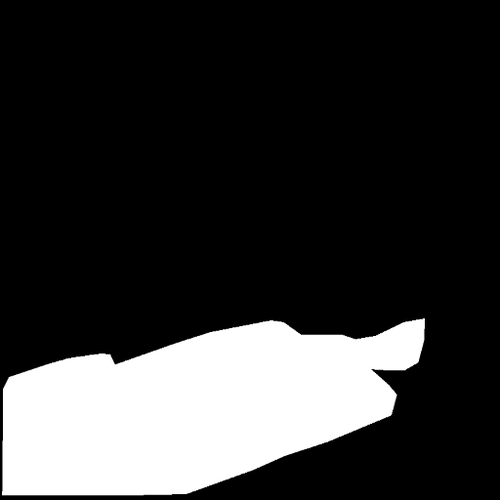} &
\includegraphics[width=\imgwidth]{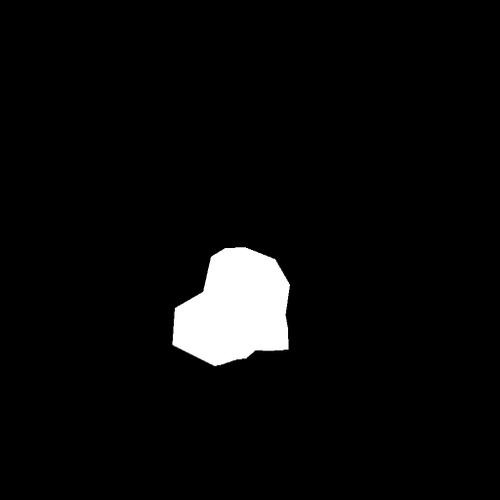} &
\includegraphics[width=\imgwidth]{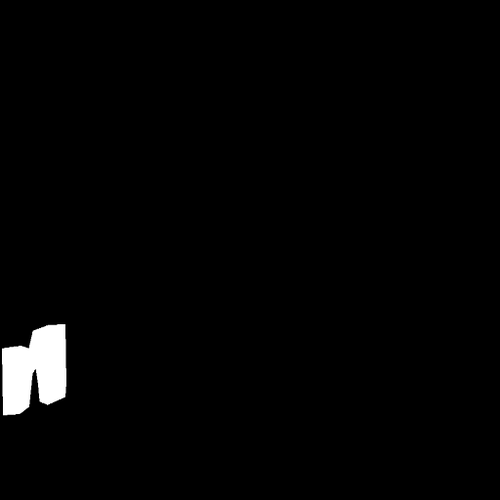} &
\includegraphics[width=\imgwidth]{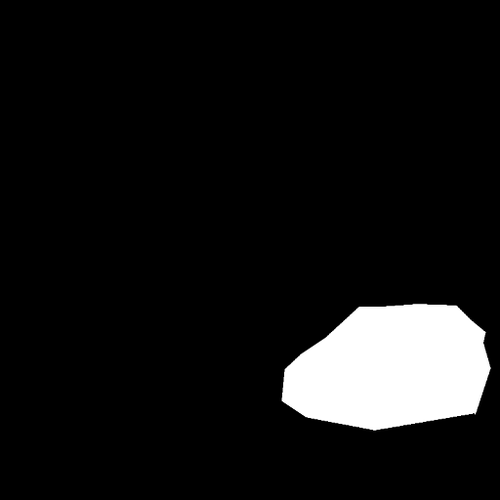} &
\includegraphics[width=\imgwidth]{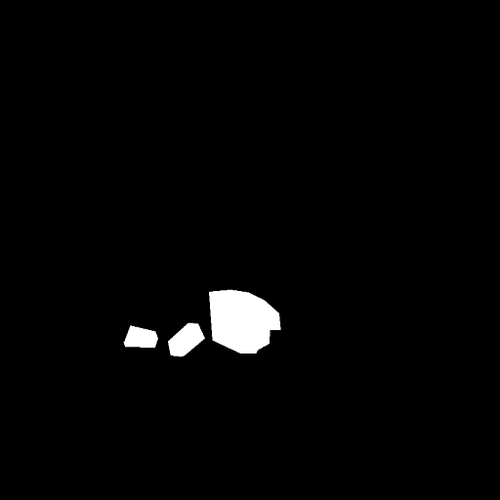} \\[-0.3em]

\multicolumn{5}{c}{Ground truth} \\
\end{tabular}

\caption{Qualitative examples of scene change detection results.}
\label{fig:qualitative}
\end{figure}

Qualitative examples of scene change detection results are shown in \cref{fig:qualitative}. The proposed \ac{name} is able to localize meaningful changes across different scenes despite variations in viewpoint and illumination. As illustrated, the generated change heatmaps highlight regions corresponding to structural or semantic changes while remaining relatively robust to noisy variations. These results further demonstrate the effectiveness of the patch-wise hashing formulation in capturing spatially localized changes using compact hash representations.

\section{Conclusion}
\label{sec:conclusion}

In this paper, we introduced \ac{name}, the first hashing-based framework for scene change detection that jointly supports global comparison and localized change identification. By constructing patch-wise hash embeddings and aggregating them through an XOR-like operation, the proposed method bridges the efficiency of global hashing with the spatial awareness required for change localization. An unsupervised contrastive learning objective enables effective training in the Hamming space without requiring pixel-level annotations.

Extensive experiments on scene change detection and retrieval benchmarks demonstrate that \ac{name} achieves competitive performance compared to state-of-the-art methods, while offering significant advantages in computational efficiency and storage. In particular, the hash-based formulation allows retrieval and change detection without re-processing previously observed images, making the method well-suited for long-term and large-scale monitoring scenarios. However, in more complex scenarios involving viewpoint shifts, camera motion, or imperfect alignment, existing methods remain limited, indicating that robust scene change detection under such conditions still requires further research. Future work will focus on extending the framework to multi-temporal sequences and viewpoint misalignment for more complex scene dynamics.

\subsubsection{Acknowledgements} This work benefited from access to the computing resources of the L3i laboratory, operated and hosted by the La Rochelle University. It is financed by the ANR (French national research agency) project ExcelLR referenced under ANR-21-EXES-0010.

\bibliographystyle{splncs04}
\bibliography{reference}

\clearpage
\setcounter{page}{1}

\title{From Image Hashing to Scene Change Detection (Supplementary Material)}

\author{Anh-Kiet Duong\inst{1, 2}\orcidID{0009-0005-0230-6104} \and
Marie-Claire Iatrides\inst{1,3}\orcidID{0009-0005-3961-0564} \and
Petra Gomez-Krämer\inst{1}\orcidID{0000-0002-5515-7828} \and
Jean-Michel Carozza\inst{2}\orcidID{0000-0001-9077-9182}}
\authorrunning{Duong et al.}

\institute{L3i Laboratory, La Rochelle University, 17042 La Rochelle Cedex 1 - France\and
LIENSs Laboratory, La Rochelle University, 17042 La Rochelle Cedex 1 - France\and
Association Ferrocampus, 17100 Saintes - France
\email{\{anh.duong,marie-claire.iatrides,petra.gomez,jean-michel.carozza\}@univ-lr.fr}}

\maketitle

In the supplementary materials, we provide additional details
and analyses to complement the main paper. In \cref{sec:proof} we provide a proof of the theoretical properties of our proposed method. In \cref{sec:addexp} we present supplementary experiments, including the data augmentation techniques used during training, efficiency evaluation on GPU and CPU, and change detection performance in terms of Intersection over Union (IoU) on the VL-CMU-CD and PCD datasets.

\section{Mathematical proof}
\label{sec:proof}

In this section, we analyze the recursive absolute-difference aggregation defined in \cref{eq:xor}, focusing on its order invariance and its approximation to the XOR operation. 

As commonly adopted in deep hashing, soft hash codes produced by a saturating activation converge arbitrarily close to binary values after training, as formally analyzed in~\cite{cao2017hashnet}. Accordingly, for each patch-wise hash $h_i \in [0,1]^l$, we assume the existence of a binary code $H_i \in \{0,1\}^l$ such that
\[
\|h_i - H_i\| \le \varepsilon,
\]
where $\varepsilon$ can be made arbitrarily small.

We first consider the aggregation of three elements, which already captures the behavior of the recursive formulation. 
In the binary domain, for $H_1,H_2,H_3 \in \{0,1\}$, the identity $|H_i - H_j| = H_i \oplus H_j$ holds element-wise. 
Therefore, the aggregation
\[
|H_3 - |H_2 - H_1|| = H_3 \oplus (H_2 \oplus H_1)
\]
is exactly equivalent to a cascaded XOR operation. 
Due to the associativity and commutativity of XOR, the result is invariant to permutations of the indices, establishing exact order invariance in the binary domain.

We now analyze the continuous case. 
Using the fact that the absolute value is a 1-Lipschitz function, we obtain
\begin{align*}
&\big| |h_3 - |h_2-h_1|| - (H_3 \oplus (H_2 \oplus H_1)) \big| \\
&= \big| |h_3 - |h_2-h_1|| - |H_3 - |H_2-H_1|| \big| \\
&\le \big| (h_3 - |h_2-h_1|) - (H_3 - |H_2-H_1|) \big| \\
&\le |h_3 - H_3| + \big| |h_2-h_1| - |H_2-H_1| \big| \\
&\le |h_3 - H_3| + |(h_2-h_1) - (H_2-H_1)| \\
&\le |h_3 - H_3| + |h_2 - H_2| + |h_1 - H_1| \\
&\le 3\varepsilon .
\end{align*}

This shows that the recursive absolute-difference aggregation is a continuous relaxation of the XOR operation, with an approximation error bounded by the soft-to-hard binarization gap. 

Let $\pi$ and $\pi'$ be two arbitrary permutations of $\{1,2,3\}$, and define
\[
A_\pi(h_1,h_2,h_3) := |h_{\pi(3)} - |h_{\pi(2)} - h_{\pi(1)}|| .
\]
Then, for any $h_i \in [0,1]$ satisfying $\|h_i - H_i\| \le \varepsilon$, we have
\[
\big| A_\pi(h_1,h_2,h_3) - A_{\pi'}(h_1,h_2,h_3) \big|
\le 6\varepsilon .
\]

By the triangle inequality,
\[
\big| A_\pi(h) - A_{\pi'}(h) \big|
\le \big| A_\pi(h) - A_\pi(H) \big|
+ \big| A_{\pi'}(h) - A_{\pi'}(H) \big|.
\]
Since $A_\pi(H) = A_{\pi'}(H)$ due to the commutativity and associativity of XOR,
and each term is bounded by $3\varepsilon$ as shown above, the result follows.
\hfill$\square$

The same bound holds for any permutation of $(h_1,h_2,h_3)$, implying approximate order invariance in the continuous domain.

Since the aggregation in \cref{eq:xor} is obtained by recursively applying the same operator over all patch-wise hashes, the above argument for three elements directly extends to an arbitrary number of patches by induction. 

Consequently, the proposed aggregation is exactly equivalent to XOR and order-invariant in the binary domain, while remaining an order-invariant approximation of XOR in the continuous domain, with a bounded and diminishing approximation error.

\section{Supplementary experiments}
\label{sec:addexp}
In this section, we provide additional experimental results to complement the main paper. We detail the data augmentation and evaluation protocol as follows. During training, we employ a diverse set of augmentations including \textit{Rotate}, \textit{RandomResizedCrop}, \textit{ColorJitter}, \textit{RandomRain}, \textit{RandomShadow}, \textit{SaltAndPepper}, \textit{RandomSnow}, \textit{ElasticTransform}, \textit{GaussNoise}, \textit{GaussianBlur}, and \textit{ToGray} to enhance robustness against appearance and environmental variations. For efficiency evaluation, inference is conducted on both GPU and CPU. On an NVIDIA H100 GPU, the average inference time is $17.6\mathrm{ms} \pm 30.8\mu\mathrm{s}$, while on CPU it is $1.09\mathrm{s} \pm 75.5\mathrm{ms}$. In \cref{tab:change_iou} we report the change detection performance in terms of Intersection over Union (IoU) on the VL-CMU-CD and PCD datasets. 

\begin{table}[ht]
\centering
\caption{Change detection performance (IoU) on VL-CMU-CD and PCD datasets.}
\label{tab:change_iou}
\begin{tabular}{lccc}
\hline
Method                                & Bits       & VL-CMU-CD      & PCD            \\ \hline
\multicolumn{4}{l}{\textit{Supervised}}                                              \\
\multicolumn{2}{l}{C-3PO \cite{wang2023reduce}}    & 0.659          & 0.700          \\ \hline
\multicolumn{4}{l}{\textit{Zero-shot}}                                               \\
\multicolumn{2}{l}{ZSSCD \cite{cho2025zero}}       & 0.348          & 0.394          \\ \hline
\multicolumn{4}{l}{\textit{Unsupervised (continuous features)}}                      \\
\multicolumn{2}{l}{SSCD \cite{ramkumar2021self}}   & 0.593          & 0.411          \\
\multicolumn{2}{l}{D-SSCD \cite{ramkumar2021self}} & 0.569          & 0.472          \\ \hline
\multicolumn{4}{l}{\textit{Unsupervised (hashing)}}                                  \\
\multirow{3}{*}{Our \ac{name}}        & 16         & 0.472          & 0.466          \\
                                      & 32         & 0.516          & 0.479          \\
                                      & 64         & \textbf{0.531} & \textbf{0.533} \\ \hline
\end{tabular}
\end{table}

\end{document}